\documentclass{article}

\usepackage[final]{neurips_2022}

\usepackage{graphicx}
\usepackage{color, colortbl}
\usepackage[utf8]{inputenc} 
\usepackage[T1]{fontenc}    
\usepackage{hyperref}       
\usepackage{url}            
\usepackage{booktabs}       
\usepackage{nicefrac}       
\usepackage{microtype}      
\usepackage{xcolor}         
\usepackage{algorithmic}
\usepackage{graphicx}
\usepackage[linesnumbered,ruled]{algorithm2e}
\usepackage{empheq}
\usepackage{listings}
\usepackage{pythonhighlightedited}
\usepackage{cancel}
\usepackage{soul}
\usepackage[most]{tcolorbox}
\usepackage{enumerate}
\usepackage{caption}

\usepackage{amsmath,amsfonts,bm}

\def\eqref#1{equation~\ref{#1}}

\def\1{\bm{1}}

\DeclareMathAlphabet{\mathsfit}{\encodingdefault}{\sfdefault}{m}{sl}
\SetMathAlphabet{\mathsfit}{bold}{\encodingdefault}{\sfdefault}{bx}{n}

\usepackage[noabbrev,capitalize]{cleveref}

\crefname{equation}{equation}{equations}
\crefname{line}{line}{lines}
\crefname{section}{\S}{\S\S}
\Crefformat{section}{#2\S#1#3}
\crefrangeformat{section}{\S\S#3#1#4--#5#2#6}

\usepackage{color}

\crefname{section}{§}{§§}
\title{TaskMatrix.AI: Completing Tasks by Connecting Foundation Models with Millions of APIs}

\author{
    Yaobo Liang\thanks{equal contribution},~~Chenfei Wu$^{*}$,~~Ting Song$^{*}$,~~Wenshan Wu$^{*}$,~~Yan Xia,~~Yu Liu,~~Yang Ou,\\
    ~~\textbf{Shuai Lu},~~\textbf{Lei Ji},~~\textbf{Shaoguang Mao},~~\textbf{Yun Wang},~~\textbf{Linjun Shou},~~\textbf{Ming Gong},~~\textbf{Nan Duan}\thanks {corresponding author}\\
    Microsoft \\
    \texttt{\{yalia, chewu, tsong, wenswu, yanxia, yluiu, yang.ou, shuailu, leiji,} \\
    \texttt{shaoguang.mao, wangyun, lisho, migon, nanduan\}@microsoft.com}\\
}

\begin{document}

\maketitle

\begin{abstract}

Artificial Intelligence (AI) has made incredible progress recently. On the one hand, advanced foundation models like ChatGPT can offer powerful conversation, in-context learning and code generation abilities on a broad range of open-domain tasks. They can also generate high-level solution outlines for domain-specific tasks based on the common sense knowledge they have acquired. However, they still face difficulties with some specialized tasks because they lack enough domain-specific data during pre-training or they often have errors in their neural network computations on those tasks that need accurate executions. On the other hand, there are also many existing models and systems (symbolic-based or neural-based) that can do some domain-specific tasks very well. However, due to the different implementation or working mechanisms, they are not easily accessible or compatible with foundation models. Therefore, there is a clear and pressing need for a mechanism that can leverage foundation models to propose task solution outlines and then automatically match some of the sub-tasks in the outlines to the off-the-shelf models and systems with special functionalities to complete them. Inspired by this, we introduce TaskMatrix.AI as a new AI ecosystem that connects foundation models with millions of APIs for task completion. Unlike most previous work that aimed to improve a single AI model, TaskMatrix.AI focuses more on using existing foundation models (as a brain-like central system) and APIs of other AI models and systems (as sub-task solvers) to achieve diversified tasks in both digital and physical domains. As a position paper, we will present our vision of how to build such an ecosystem, explain each key component, and use study cases to illustrate both the feasibility of this vision and the main challenges we need to address next.

\end{abstract}

\section{Introduction} 

\begin{center} ``\textit{The amount of intelligence in the universe doubles every 18 months.'' ~-- ~Sam Altman, OpenAI CEO} \end{center}

Foundation models have made remarkable progress in this decade, from understanding models (e.g., BERT~\citep{devlin2018bert}, ViT~\citep{DBLP:conf/iclr/DosovitskiyB0WZ21vit}, Whisper~\citep{radford2022whisper}) that can process and comprehend data of different modalities, to generative models (e.g., GPT-4~\citep{openai2023gpt4}, GPT-3~\citep{brown2020gpt3}, Codex~\citep{chen2021codex}, DALL·E~\citep{ramesh2021dalle}) that can produce various kinds of outputs to interact with the world. ChatGPT is so impressive that many people think it is a sign of Artificial General Intelligence (AGI) coming soon. However, foundation models still face limitations and challenges in doing some specialized tasks, such as performing accurate mathematical calculations or completing a multi-step task in the real world that requires both textual and visual processing skills. Meanwhile, there are existing models and systems (based on mathematical theories, symbolic rules, or neural networks) that can perform very well on some domain-specific tasks. But due to the different implementation or working mechanisms, they are not readily available or compatible with foundation models. Therefore, there is an urgent and obvious need for a mechanism that can link the foundation models with the off-the-shelf models and systems with special functionalities to finish diversified tasks in both the digital and physical worlds.

Motivated by this, we present our vision of building a new AI ecosystem, named as \textbf{TaskMatrix.AI}, for linking foundation models with millions of existing models and system APIs to finish diversified tasks. Different from any single AI model, TaskMatrix.AI can be seen as a super-AI with abilities to execute both digital and physical tasks, which has the following key advantages:

\begin{itemize}
\item{\textbf{TaskMatrix.AI can perform both digital and physical tasks} by using the foundation model as a core system to understand different types of inputs (such as text, image, video, audio, and code) first and then generate codes that can call APIs for task completion.}
\item{\textbf{TaskMatrix.AI has an API platform as a repository of various task experts}. All the APIs on this platform have a consistent documentation format that makes them easy for the foundation model to use and for developers to add new ones.} 
\item{\textbf{TaskMatrix.AI has a powerful lifelong learning ability}, as it can expand its skills to deal with new tasks by adding new APIs with specific functions to the API platform.} 
\item{\textbf{TaskMatrix.AI has better interpretability for its responses}, as both the task-solving logic (i.e., action codes) and the outcomes of the APIs are understandable.}

\end{itemize}

Our vision is to build an ecosystem that can leverage both foundation models and other models and systems that are good at specific tasks and can be accessed as APIs. By connecting the foundation model with APIs, TaskMatrix.AI can smoothly integrate neural and symbolic systems, accomplish digital and physical tasks, and provide strong lifelong learning and reliable capabilities.
\section{TaskMatrix.AI Architecture}

\begin{figure*}[t]
    \centering
    \includegraphics[width=\textwidth]{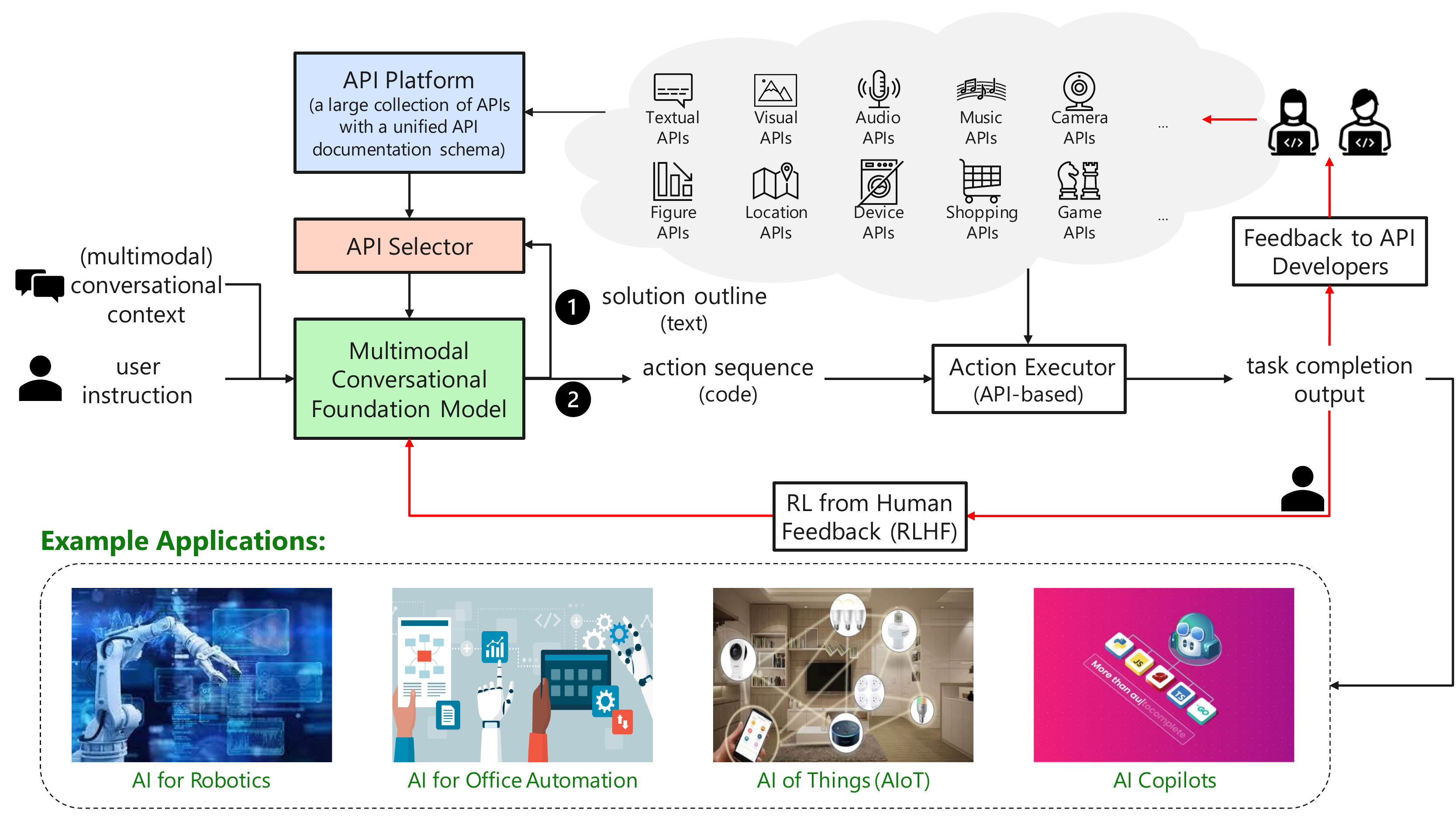}
    \caption{Overview of TaskMatrix.AI. Given user instruction and the conversational context, the multimodal conversational foundation model (MCFM) first generates a solution outline (step \textcircled{1}), which is a textual description of the steps needed to solve the task. Then, the API selector chooses the most relevant APIs from the API platform according to the solution outline (step \textcircled{2}). Next, MCFM generates action codes using the recommended APIs, which will be further executed by calling APIs. Last, the user feedback on task completion is returned to MCFM and API developers.}
    \label{fig:overview}
\end{figure*}

\subsection{Overview}

The overall architecture (Figure 1) of TaskMatrix.AI consists of the following four key components: (1) \textbf{Multimodal Conversational Foundation Model (MCFM)}, which is responsible for communicating with users, understanding  their goals and (multimodal) contexts, and generating executable codes based on APIs to accomplish specific tasks. (2) \textbf{API Platform}, which provides a unified API documentation schema to store millions of APIs with different kinds of functions and allows API developers or owners to register, update and delete their APIs. (3) \textbf{API Selector}, which can recommend related APIs based on MCFM’s comprehension of the user command. (4) \textbf{API Executor}, which can execute the generated action codes by calling the relevant APIs and return the intermediate and final execution results. 

The key procedure in this architecture is the MCFM's ability to generate action codes based on user instructions. We formulate this procedure as:
\begin{equation}
    \mathcal{A}=\mathrm{MCFM}(\theta, \mathcal{P}, \mathcal{I}, \mathcal{C})
\end{equation}
The MCFM takes four inputs: the parameter of the foundation model, denoted as $\theta$; the API platform, denoted as $\mathcal{P}$; the user instruction, denoted as $\mathcal{I}$; and the conversational context, denoted as $\mathcal{C}$. Using these inputs, the model generates action codes, denoted as $\mathcal{A}$, to accomplish the user's instruction. 

TaskMatrix.AI also provides two learnable mechanisms to align MCFM with APIs in a better way. Both these two mechanisms need user feedback, denoted as $\mathcal{F}$. We calculate the loss function, denoted as $\mathcal{L}(\mathcal{A}, \mathcal{F})=\mathcal{L}(\mathrm{MCFM}(\theta, \mathcal{P}, \mathcal{I}, \mathcal{C}), \mathcal{F})$. This loss function can then be optimized by tuning the parameter of the foundation model, $\theta$, and updating the document in the API platform, $\mathcal{P}$. First, the output signals can be used by the \textbf{Reinforce Learning from Human Feedback (RLHF)} mechanism to enhance MCFM’s skill in API comprehension and action code generation from user commands, as well as the retrieval performance of API selector. We represent it as $min_{\theta} \mathcal{L}(\mathrm{MCFM}(\theta, \mathcal{P}, \mathcal{I}, \mathcal{C}), \mathcal{F})$. Second, the output signals can be also used as the \textbf{Feedback to API Developers}, for them to improve the documentation of APIs and make them easier to understand and called by the MCFM. We represent it as $min_{\mathcal{P}} \mathcal{L}(\mathrm{MCFM}(\theta, \mathcal{P}, \mathcal{I}, \mathcal{C}), \mathcal{F})$.

\subsection{Multimodal Conversational Foundation Model (MCFM)}

An ideal MCFM should have four main capabilities: (1) It should be able to take multimodal inputs and contexts (such as text, image, video, audio, and code) and generate executable codes based on APIs that can complete specific tasks. Most existing multimodal models (e.g., CLIP and Flamingo) are not suitable for this task as they can only encode different modalities but lack the conversational ability and code-generation skills. ChatGPT is a model that can understand language and conversation well and generate code accordingly, but it only works with text and code modalities. GPT-4 is the most suitable model until now, as it can deal with multimodal inputs and generate both text and code as outputs. (2) It should be able to extract specific tasks from user instructions and propose reasonable solution outlines (as shown in Figure 1) that can help select the most relevant APIs for code generation. Both ChatGPT and GPT-4 have this capability as they were pre-trained on both text and code corpora, which gives these two models strong knowledge to reason and plan. (3) It should be able to quickly learn how to use APIs from their documentation and match them to specific tasks based on common sense and API usage history. (4) It should incorporate an explicit code verification mechanism to confirm the reliability and trustworthiness of the generated codes.

With these capabilities, MCFM is involved in two primary steps (illustrated as step \textcircled{1} and step \textcircled{2} in Figure 1). First, step \textcircled{1} takes each user instruction and the corresponding conversational context as input and generates a solution outline. Users often use brief expressions to convey their high-level task intentions, so MCFM generates a more comprehensive textual description of the steps required to complete the task by leveraging its deep understanding of world knowledge. Users can then actively edit the outline during the conversation. MCFM can also edit the outline in cases where there is no suitable API to fulfill a particular step, or where the result of a certain step fails to satisfy the user's instructions. However, if the user's instructions already provide sufficient details for task completion, this step can be skipped. Second, after the API selector takes the solution outline as input and retrieves related APIs, step \textcircled{2} generates the code of action using the selected APIs. Here, MCFM must support acting with a dynamic action space, as developers continuously upload and modify APIs, and the retrieval results vary for different user instructions. While generating a sequence of actions is often sufficient, incorporating action codes as generation results can enhance the expression capacity.

\subsection{API Platform}
\label{sec:overview_api_platform}

The API platform has two main functions: first, it provides storage for different types of APIs that MCFM can access; second, it allows API developers or owners to manage their APIs by registering, updating, or deleting them.
To help MCFM understand and utilize APIs better, the API platform specifies a unified API documentation schema, which consists of five aspects for each API document:

\begin{itemize}
    \item \textbf{API Name}: The API name provides an abstract of the API. It helps MCFM to link user instructions to this API and serves as an entry for the action executor. The name should be clear and precise in natural language and avoid ambiguity with other API names. 
    \item \textbf{Parameter List}: The parameter list for an API includes the input parameters and return value, and each parameter has a parameter name, parameter description, data type, and default value. This information can assist MCFM in correctly filling the parameters in the corresponding positions with the appropriate format.
    \item \textbf{API Description}: Compared to the API name, the API description contains more information about what the API does, how it works, what are its inputs and outputs, and any potential errors or exceptions that may be raised.
    \item \textbf{Usage Example (Optional)}: Providing a usage example for complex APIs can help demonstrate how the API can be used, while it may not be necessary for simple APIs.
    \item \textbf{Composition Instructions (Optional)}: Developers who offer a package of APIs could provide composition instructions. This can serve as guidance to the model on how to combine multiple APIs to accomplish complex user instructions. For instance, in a file editing scenario, the model may need to open the target file before making edits and then save the file after completing the edits.
\end{itemize}
We provide an example of an API document for opening a file, which is a simplified version of $open$ in python.\\

\noindent\fbox{%
  \parbox{0.985\textwidth}{%
\textbf{API Name}: \pyth{open\_local\_file} \\
\textbf{API Parameter}: \pyth{(file_path:string, model: string="r")}. \pyth{file_path: string}, the pathname (absolute or relative to the current working directory) of the file to be opened. \pyth{mode: string="r"}, the mode is an optional string that specifies the mode in which the file is opened. It defaults to \pyth{"r"} which means open for reading in text mode. Other common values are \pyth{"w"} for writing. This file will return a File object or OSError. \\
\textbf{API Description}: Open the file and return a corresponding file object. If the file cannot be opened, an OSError is raised.\\
\textbf{Usage Example}: \pyth{f = open\_local\_file("example.txt", "w")} \\
\textbf{Composition Instructions}: Open should be used before reading and editing. The file should be closed by \pyth{close_local_file} after all operations.
  }
}

\subsection{API Selector}

The goal of API selector is to identify and select the most suitable APIs from API platform that fit the task requirement and solution outline as understood by MCFM. Since the API platform may have millions of APIs, the API selector needs the search capability to retrieve semantically relevant APIs. The API selector can also leverage a module strategy to quickly locate relevant APIs. Each package corresponds to a specific domain, such as a package of visual models, math, specific software, or specific physical devices.

\subsection{Action Executor}

The action executor is designed to execute the action codes. TaskMatrix.AI uses an action executor to run various APIs, ranging from simple HTTP requests to complex algorithms or AI models that need multiple input parameters. After the execution, the action executor will return the results to users. To enhance accuracy and reliability, the action executor also requires a verification mechanism to confirm whether the generated code or outcomes satisfy the tasks specified in human instructions.

\subsection{Reinforcement Learning with Human Feedback (RLHF)}

RLHF is a general technique that uses reinforcement learning methods to optimize machine learning models based on human feedback. It has been successfully used to align large models trained on the general corpus of text data with user instructions, such as InstructGPT \citep{ouyang2022training}.

In TaskMatrix.AI, we use RLHF to benefit from the knowledge and insight of human feedback to enhance MCFM and API selector. This can result in faster convergence and better performance of TaskMatrix.AI on complex tasks.
Specifically, we leverage human feedback to train a reward model that can classify whether the task has been completed. During RLHF training, MCFM and API selector can explore various strategies to plan solution outlines, select and compose APIs, and the reward model can provide feedback. Using RLHF, MCFM and API selector can optimize their policy and discover better ways to accomplish tasks. 

\subsection{Feedback to API Developers}
After TaskMatrix.AI has performed a specific task, the user feedback will be delivered to the API developers in an appropriate manner to indicate whether their APIs have been successfully used to complete the task or not. Such <user instruction, API calls, user feedback> triples can serve as either demonstration examples of how to use specific API, or as guidance for API developers to improve the API documentations to make them more understandable for MCFM and API selector.

Specifically, we treat the API documentation as learnable parameters, similar to the parameters of MCFM. User feedback can help the developer to understand how well the API works during different inputs and when combined with different APIs. This step can also be aided by a model, such as ChatGPT, that takes human feedback as input and generates natural language suggestions to improve the API documentation. We provide an example in Section \ref{sec:feedback_to_api_doc}.

\section{Application Scenarios}

In this section, we present some examples of how TaskMatrix.AI can be applied in different application scenarios. We show how TaskMatrix.AI can assist in creating AI-powered content in Section \ref{sec:visualchatgpt} and \ref{sec:complex_aigc}. We demonstrate how TaskMatrix.AI can facilitate office automation and cloud service usage in Section \ref{sec:ppt} and \ref{sec:cloud}. We illustrate how TaskMatrix.AI can perform tasks in the physical world by interacting with robots and IoT devices in Section \ref{sec:iot}. \textit{All these cases have been implemented in practice and will be supported by the online system of TaskMatrix.AI, which will be released soon.} We also explore more potential applications in Section \ref{sec:more}.

\subsection{Visual Task Completion}
\label{sec:visualchatgpt}

\begin{figure*}
    \centering
\includegraphics[width=1.0\textwidth]{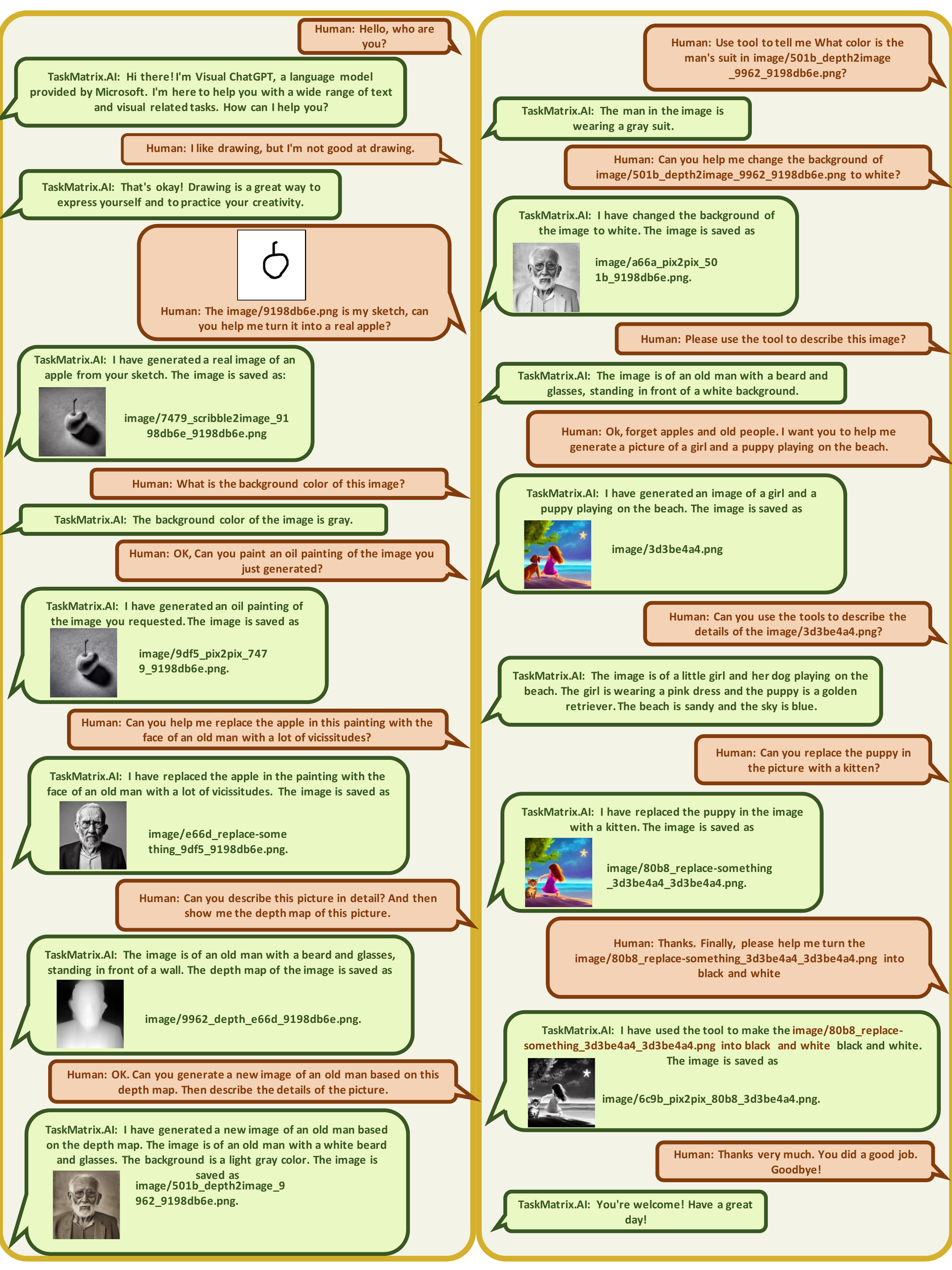} 
\caption{Multiple rounds of dialogue between human and Visual ChatGPT~\citet{wu2023visualchatgpt}. In the dialogues, Visual ChatGPT, an initial version TaskMatrix.AI, can understand human intentions, support the language and image inputs, and provide complex visual tasks such as generation, question, and editing.} 
\label{fig:FullCase}
\end{figure*}

TaskMatrix.AI enables the user to interact with AI by 1) sending and receiving not only languages but also images 2) providing complex visual questions or visual editing instructions that require the collaboration of multiple AI models with multi-steps. 3) providing feedback and asking for corrected results. We design a series of prompts to inject the visual model information into ChatGPT, considering models of multiple inputs/outputs and models that require visual feedback. More details are described at ~\citet{wu2023visualchatgpt}.
We demonstrate this with an example in Figrue~\ref{fig:FullCase}. The APIs related to this include:

\begin{figure*}
    \centering
\includegraphics[width=0.95\textwidth]{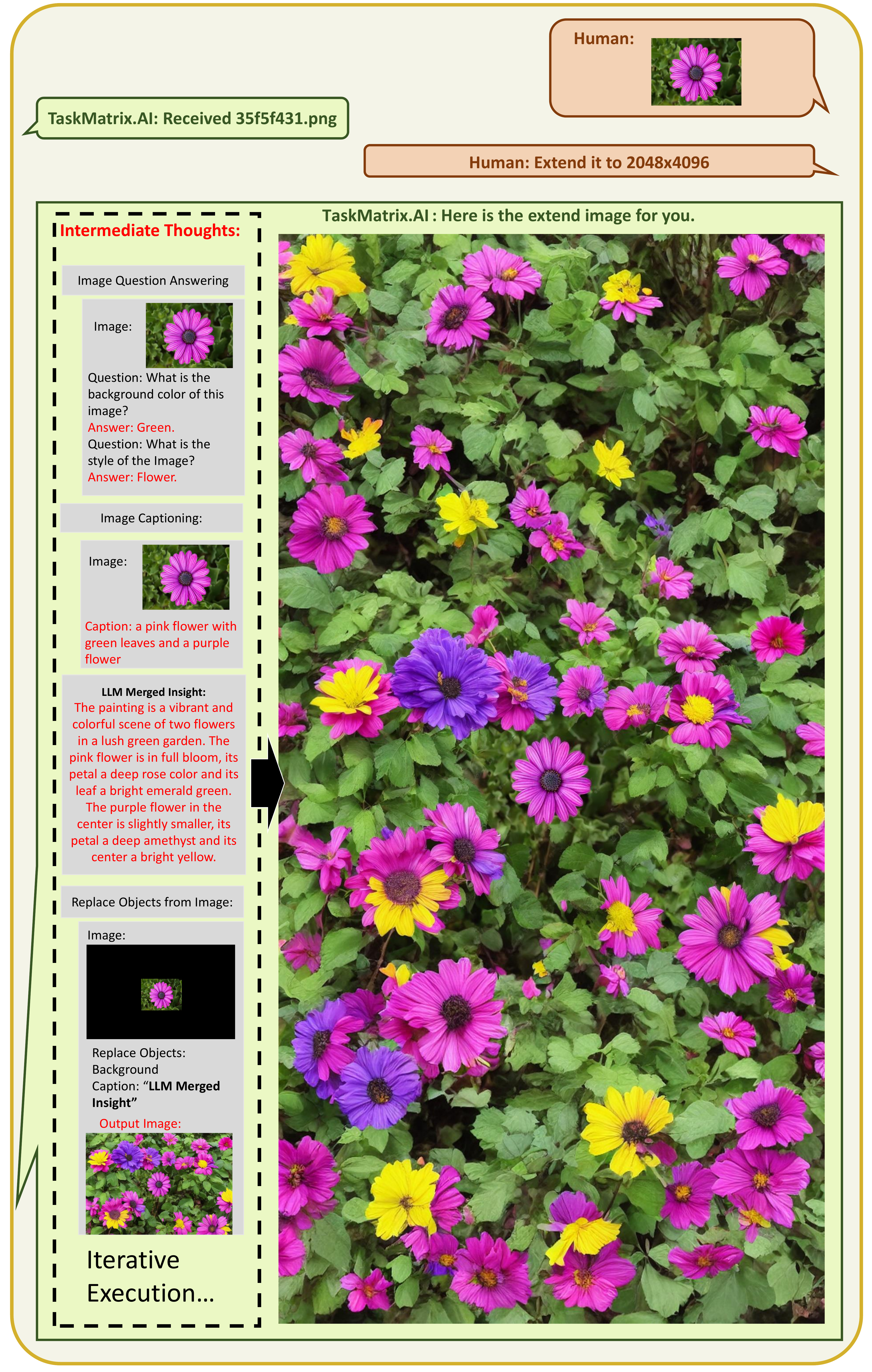} 
\caption{An image outpainting example. In this example, we define a solution outline using three APIs: Image Question Answering, Image Captioning, Replace Objects from Image. By iteratively get captions of an image and ask LLM to imagine and replace the surroundings of it, we finally get a high-resolution image of $2048\times 4096$.} 
\vspace{-5mm}
\label{fig:OutCase}
\end{figure*}

\begin{itemize}
    \item \textbf{Image Editing} Image Editing includes removing or replacing objects of an image, or changing the style of an image. Removing objects from an image involves using image editing tools or algorithms to get rid of unwanted elements. On the other hand, replacing objects with new ones involves swapping out an element in an image with another one that is more suitable. Finally, changing an image using text involves using machine learning algorithms to generate an image based on a textual description.
    \item \textbf{Image Question Answering}  This refers to the process of using machine learning algorithms to answer questions about an image, often by analyzing the contents of the image and providing relevant information. This can be useful in situations where the image contains important information that needs to be extracted.
    \item \textbf{Image Captioning} This refers to the process of using machine learning algorithms to generate textual descriptions of an image, often by analyzing the contents of the image and providing relevant information. 
    \item \textbf{Text-to-Image} This refers to the process of generating an image from a textual description, often using machine learning algorithms that can generate realistic images based on textual input. 
    \item \textbf{Image-to-Sketch/Depth/Hed/Line}  This refers to the process of converting an image to a sketch, depth, Hed (Holistically-nested edge detection), or line, often using image processing techniques or computer algorithms.
    \item \textbf{Sketch/Depth/Hed/Line-to-Image} This refers to the process of generating an image from a sketch, depth, Hed (Holistically-nested edge detection), or line.

\end{itemize}
In the Fig.~\ref{fig:FullCase} example, TaskMatrix.AI is capable of understanding images as inputs and generating images, while the MCFM in this implementation is ChatGPT, which can only process text. It demonstrates the ability to extend to more modalities without retraining the MCFM. Additionally, TaskMatrix.AI can compose multiple APIs to accomplish complex user intentions, including tasks such as generation, questioning, and editing. This approach provides more interpretability and controllability than an end-to-end model. 

Fig.~\ref{fig:OutCase} illustrates an example of high-resolution image generation, where multiple APIs collaborate to produce the final result. In this example, we define a solution outline consisting of three APIs: Image Question Answering, Image Captioning, and Replace Objects from Image. The left dashed box in Fig.~\ref{fig:OutCase} demonstrates how the solution outline assists in extending an image to a $2048\times 4096$ resolution. First, the Image Question Answering API is employed to identify two crucial features of an image: the background color and the style. These answers are essential for extension, as the expanded image should maintain the background color and style of the original image. Second, the image captioning model is utilized to obtain a description of the image, which provides fundamental information about the original image. Third, a multi-modal conversational foundation model is used to merge all the obtained information and envision the surrounding descriptions of the image. Fourth, an Image Editing API (Replace Objects from Image) is employed to substitute the surrounding unknown regions with the envisioned descriptions. By iteratively executing the four pre-defined steps in the solution outline, TaskMatrix.AI can generate a high-resolution image of any desired size (in this case, $2048\times 4096$).

\subsection{Multimodal Long Content Generation}
\label{sec:complex_aigc}
\begin{figure*}
    \centering
\includegraphics[width=1.0\textwidth]{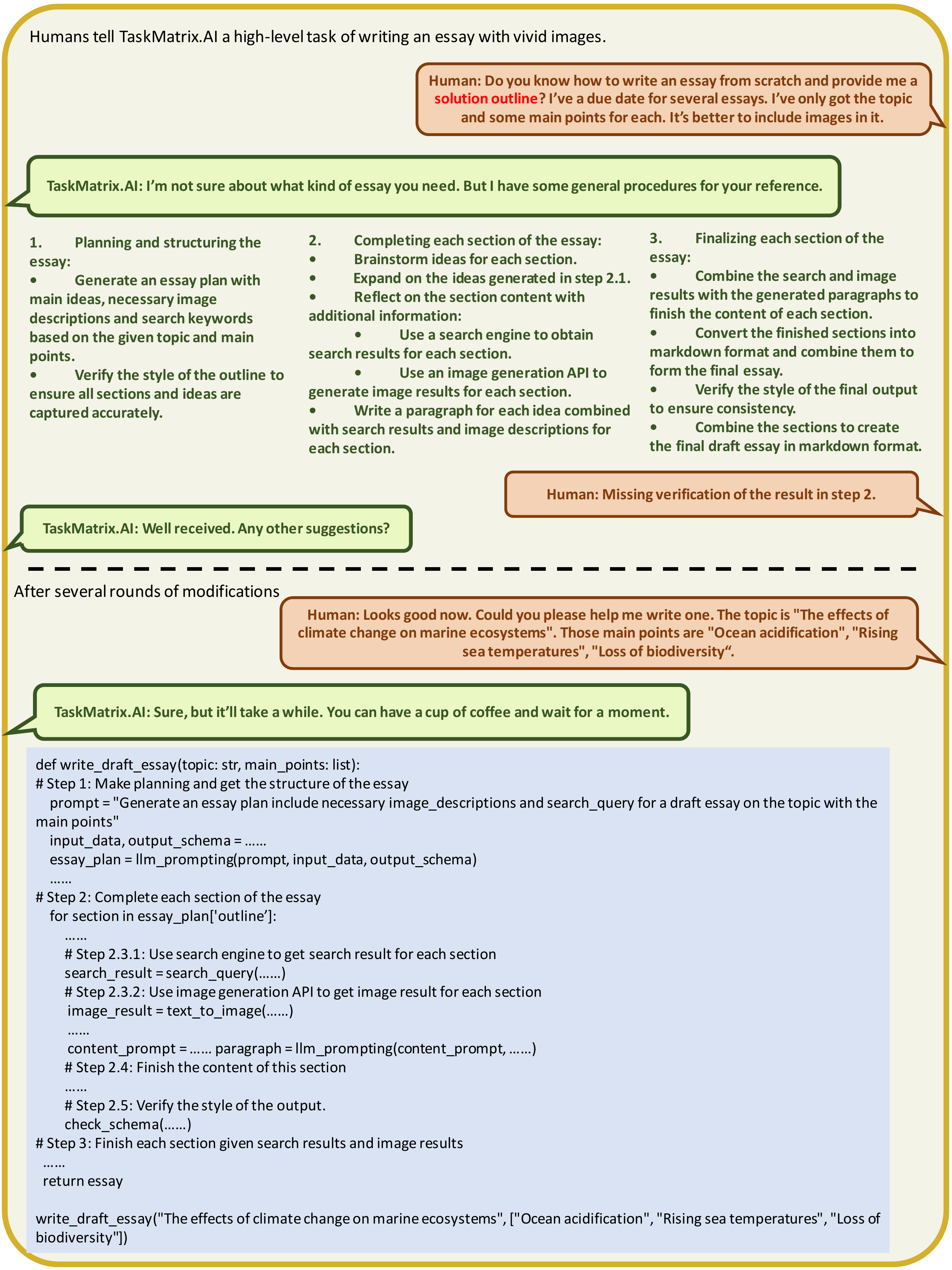} 
\caption{Multiple rounds of dialogue between user and TaskMatrix.AI. TaskMatrix.AI can comprehend high-level user instructions and assist in generating structured, multimodal content including text and image elements. TaskMatrix.AI first generates a solution outline needed to accomplish the task, makes decisions on when and how to use the APIs, then generates code based on it. The solution outline and generated codes can be applied to similar tasks of generating multimodal long content.} 
\label{fig:complex_aigc_1}
\end{figure*}

\begin{figure*}
    \centering
\includegraphics[width=1.0\textwidth]{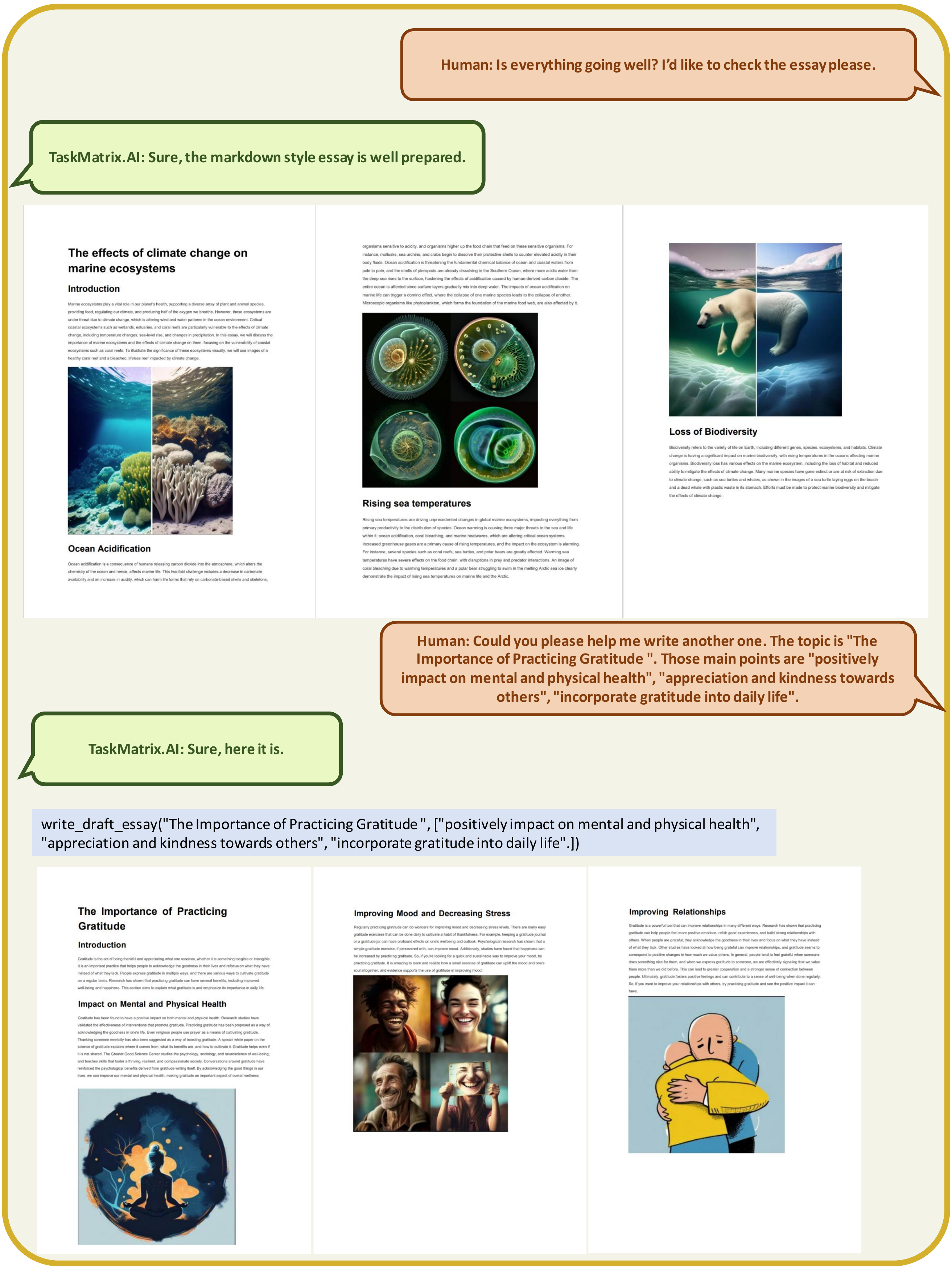} 
\caption{Multiple rounds of dialogue between user and TaskMatrix.AI. TaskMatrix.AI can comprehend high-level user instructions and assist in generating structured, multimodal content including text and image elements. TaskMatrix.AI first generates a solution outline needed to accomplish the task, makes decisions on when and how to use the APIs, then generates code based on it. The solution outline and generated codes can be applied to similar tasks of generating multimodal long content.} 
\label{fig:complex_aigc_2}
\end{figure*}

TaskMatrix.AI can help users to create multimodal long content including text and image elements. Motivated by the planning-based method in long text generation task\citep{wang2022language} aimed to improve coherence, an explicit planning process is involved to improve both textual and visual consistency in this multimodal content scenario. We show an example in Figure \ref{fig:complex_aigc_1} and Figure \ref{fig:complex_aigc_2}. TaskMatrix.AI can take high-level instructions as input, and generate a solution outline to accomplish this task. The planning enhanced MCFM automatically decomposes the task into small sub-tasks. And the solution outline in this scenario is a step-by-step proposal that could be interactively modified by the user in later conversation rounds. Based on the finalized proposal, then it can leverage API exposure and generate action codes that are capable of integrating different APIs in every single step. Finally, we can get the generated multimodal content with the action executor to run the action code. The APIs related to this example include:
\begin{itemize}
    \item \textbf{Search API} This API allows for retrieving information through a search engine. We name the API as \pyth{search_query}. Most search engines provide API for developers. 
    \item \textbf{Text-to-Image API} This API allows for generating an image from a textual description. We name the API as \pyth{text_to_image}. In our experiment, we leverage Midjourney V5\footnote{https://docs.midjourney.com/docs/models} to generate the image.
    \item \textbf{Large Language Model API} This API allows for text generation based on given prompts. We name the API as \pyth{llm_prompting}. In our experiment, we leverage the ChatGPT API from azure OpenAI service\footnote{https://learn.microsoft.com/en-us/azure/cognitive-services/openai/how-to/chatgpt} to generate text throughout the whole process.
\end{itemize}

In this example, TaskMatrix.AI can accomplish complex user instructions by planning solution outlines first, making decisions on when and how to use the APIs, then generating the action code, and finally finishing an essay including text and image elements. To be noticed, the \textbf{Large Language Model API} is fully leveraged to generate not only text elements but also search keywords for \textbf{Search API} and text prompts for \textbf{Text-to-Image API}. During the experiment, we found that this could improve the controllability and quality of generated content. Furthermore, the solution outlines can be applied to similar user instructions. Users can also reuse the solution outlines as templates. Since solution outlines are entirely described in natural language which is now becoming a general interface for large language models, they could potentially become shared knowledge among humans and models.

\subsection{Office Automation}
\label{sec:ppt}
TaskMatrix.AI has the capability to comprehend user instructions received through voice and automate the operation of the software on computers and applications on phones, as well as the operating system. People often rely on using the mouse, keyboard, and fingers to perform various tasks and have a very heavy workload in accomplishing their complex goals. TaskMatrix.AI introduces a natural language interface and automates user instructions, thereby reducing the workload. With TaskMatrix.AI, users can easily use complex software without requiring extensive training, find the right features without searching, and adapt to software updates or new software with minimal effort. This can help to relieve humans from mundane work and allow them to focus on the creative aspects of their work and make high-level decisions.

\begin{figure*}
    \centering
\includegraphics[width=1.0\textwidth]{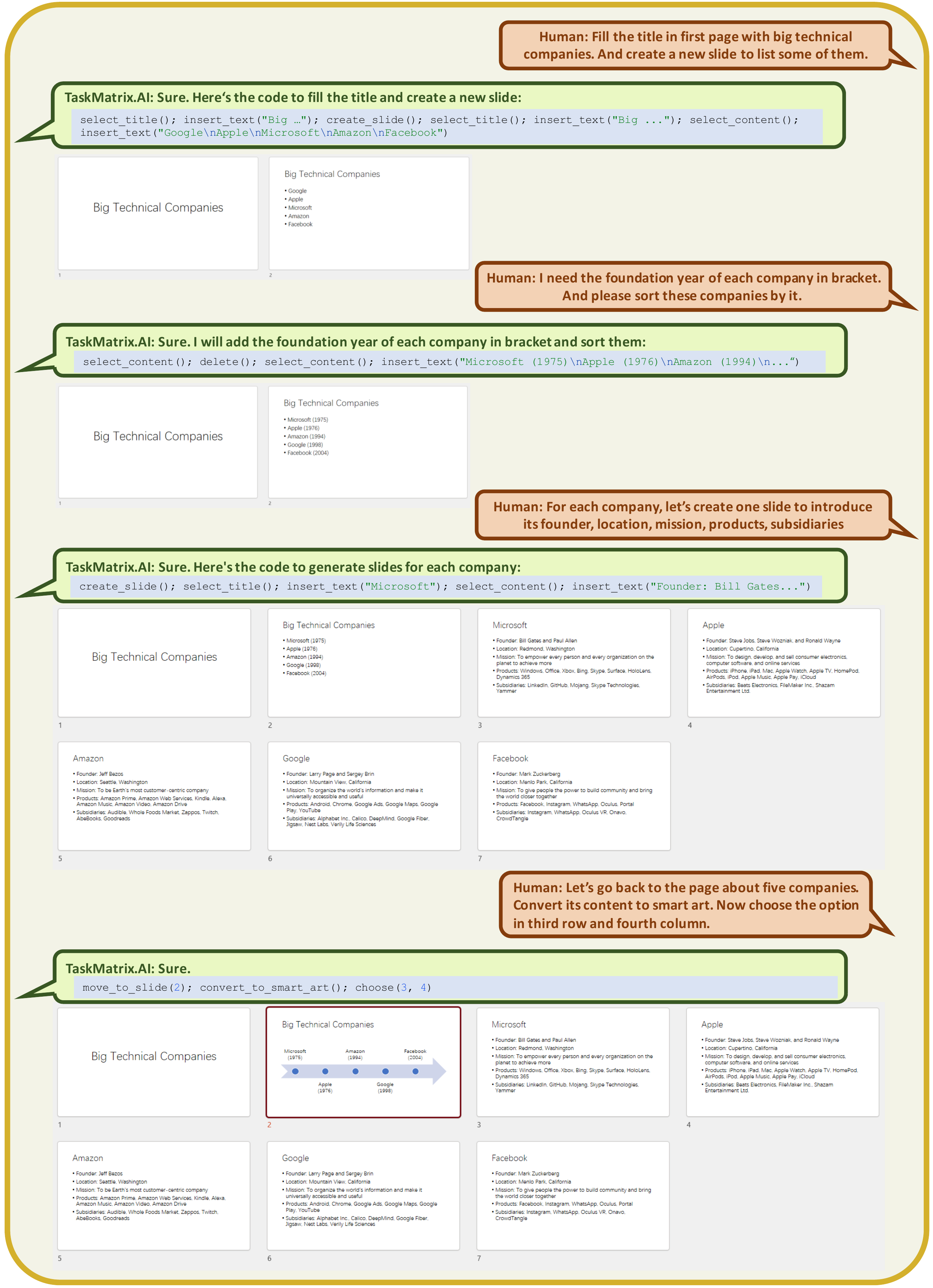} 
\caption{Multiple rounds of dialogue between user and TaskMatrix.AI. TaskMatrix.AI can understand user instructions and operate PowerPoint on behalf of users. TaskMatrix.AI is capable of breaking down the user's complex instructions into multiple PowerPoint operations, assisting users in finding and using infrequent features, and generalizing the same patterns across multiple pages. While we display the API calls in a gray text box, this information is not necessary for the user.} 
\label{fig:office_demo1}
\end{figure*}

\begin{figure*}
    \centering
\includegraphics[width=1.0\textwidth]{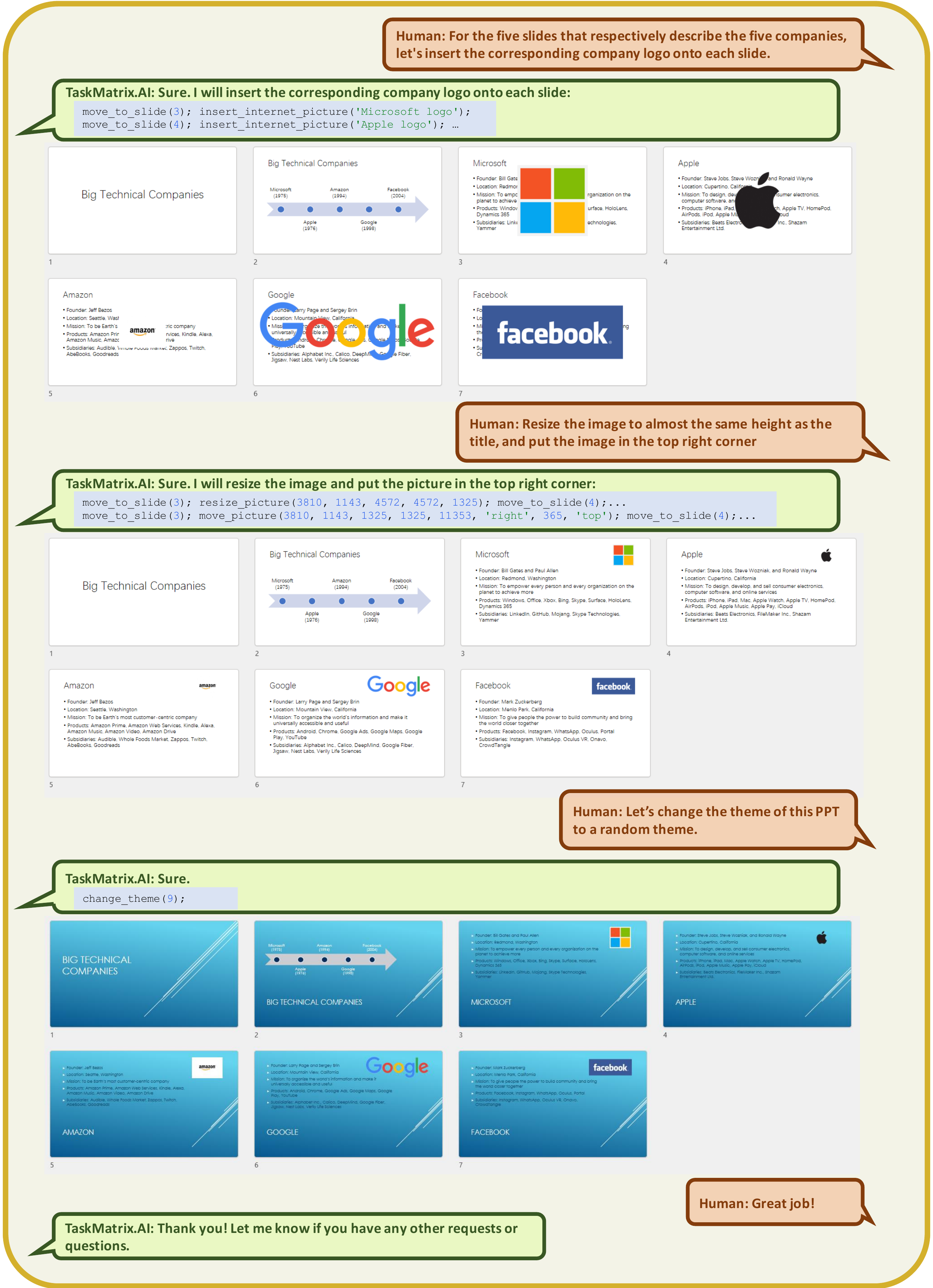} 
\caption{More rounds of dialogue between user and TaskMatrix.AI. TaskMatrix.AI can accomplish the insert logo instruction by the insert internet feature of PowerPoint with API \pyth{insert_internet_image("Microsoft logo")}. This feature will provide multiple images for users. TaskMatrix.AI can take the user's instructions to select one of them. In the example, we omitted the selection steps for brevity.} 
\label{fig:office_demo2}
\end{figure*}

We demonstrate this with an example of PowerPoint automation, shown in Figure ~\ref{fig:office_demo1} and Figure ~\ref{fig:office_demo2}. TaskMatrix.AI can help to create slides related to a specific topic, change the content, insert and adjust the images, and change themes. The details to implement this scenario is in Section \ref{sec:case_study}. The APIs related to this example include:
\begin{itemize}
    \item \textbf{Mouse and Keyboard API} To control PowerPoint, we utilize the mouse and keyboard API as it is a universal method to manipulate the operating system. This API is provided in the PyAutoGUI package\footnote{https://pyautogui.readthedocs.io/} of Python. 
    \item \textbf{PPT File Reader API} The content provides essential information to understand user instructions. We utilize the python-pptx package\footnote{https://python-pptx.readthedocs.io/} to extract content from saved PPT files. The content includes the text on each page and the position of each text box, image, and other shapes. For other software, we can replace this package with operating system APIs or visual understanding models for more flexibility.
    \item \textbf{PowerPoint APIs} We leverage the APIs provided by PowerPoint software to control it, which include the APIs to create a new slide \pyth{create_slide}, select title and content before editing it \pyth{select_title, select_content}, insert text to a specific text box \pyth{insert_text}, move to a specific page \pyth{move_to_slide}, resize and move images \pyth{resize_picture, move_picture}. We also include several infrequently used functions like converting to smart art \pyth{convert_to_smart_art}, inserting images from the internet \pyth{insert_internet_picture}, and changing themes \pyth{change_theme}.

\end{itemize}
In this example, TaskMatrix.AI is capable of decomposing high-level instructions into multiple PowerPoint APIs. For instance, the third query requires 25 APIs to complete. Additionally, TaskMatrix.AI can understand user instructions based on PowerPoint content. For instance, it can generate five pages based on the company list on page 2 and insert a logo based on the title of each page, and it can determine the page index based on the user's coarse-grained command, such as "about five companies".

\subsection{Cloud Services Utilization}
\label{sec:cloud}
TaskMatrix.AI can help users to access the services on Cloud, which provide computing, storage, networking, analytics security, and more. Cloud services offer a multitude of APIs, and new APIs are constantly being developed. TaskMatrix.AI can understand these APIs and learn new APIs, then recommend appropriate APIs based on user instructions.

\begin{figure*}
    \centering
\includegraphics[width=1.0\textwidth]{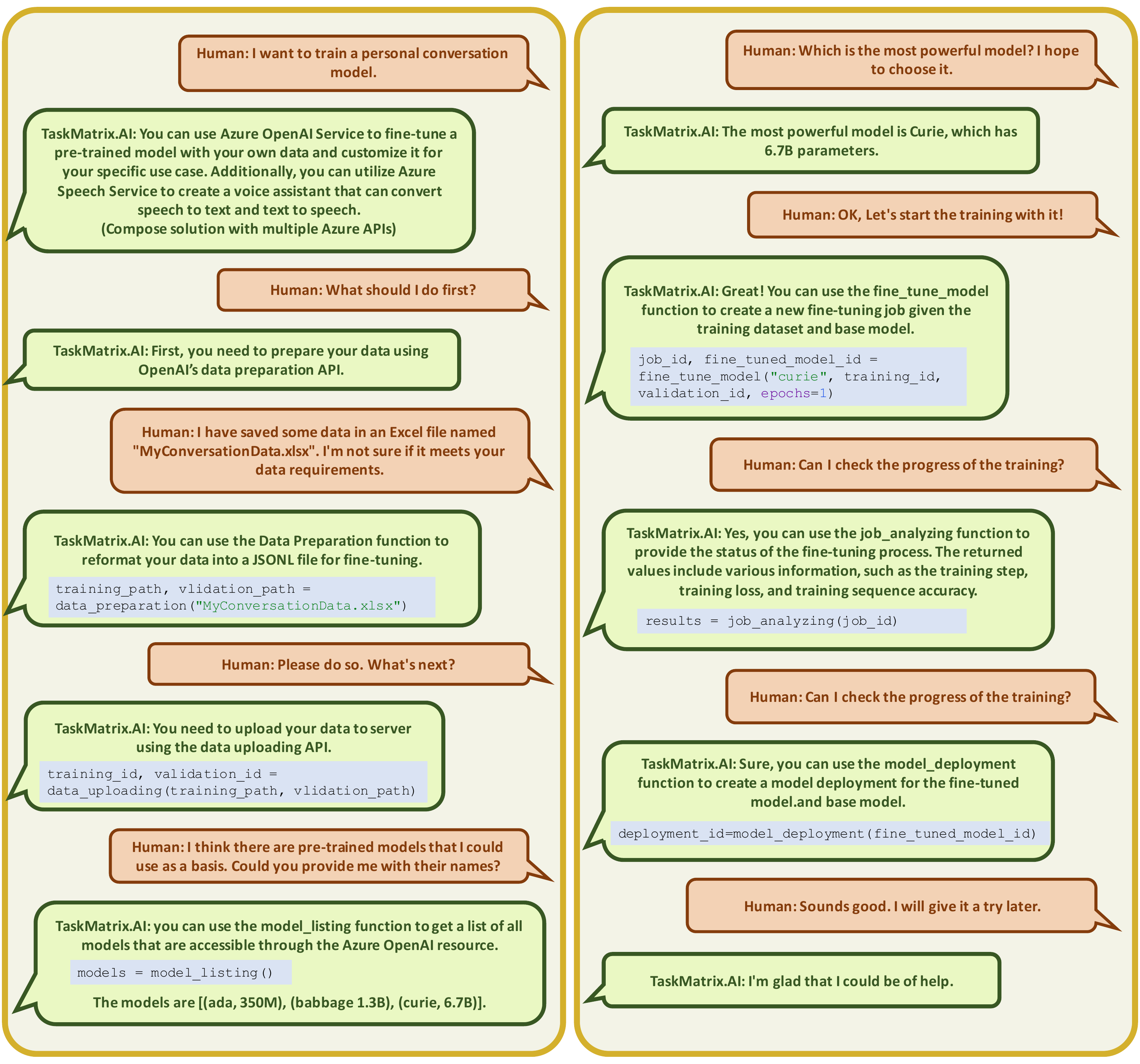} 
\caption{TaskMatrix.AI assists users in building a personal conversation model using Azure Cloud APIs. TaskMatrix.AI can compose a solution with multiple Azure APIs to accomplish user instruction. It can help user to manage data, train models, check training status, and deploy models.} 
\label{fig:azure_assistant}
\end{figure*}

In Figure \ref{fig:azure_assistant}, we provide an example of how Azure Cloud APIs can assist users in building a personal conversation model. With this example, users can easily manage data, computing resources, model training, and deployment, even with minimal expertise. This scenario was initially conceived by New Bing, and most of the step-by-step knowledge about fine-tuning the model comes from the Azure OpenAI fine-tuning document \footnote{https://learn.microsoft.com/en-us/azure/cognitive-services/openai/how-to/fine-tuning\label{azure_open_ai}}. We define the APIs based on the related documents the test whether TaskMatrix.AI can link user instructions to the correct API and fill the parameters accurately. The APIs related to this example include:
\begin{itemize}
    \item \textbf{OpenAI's Data Preparation API \textsuperscript{\ref{azure_open_ai}}}: OpenAI's CLI (command-line interface) data preparation tools can validate, give suggestions, and reformat user's data into a JSONL file for fine-tuning. This tool accepts input in various formats such as Microsoft Excel workbooks, comma-separated values, JSON Lines, and others. We name this API as \pyth{data_preparation}.
    \item \textbf{Data Uploading API\textsuperscript{\ref{azure_open_ai}}}: Azure provides APIs that enable users to upload training data to the service from either a local file or Azure Blob Storage. Alternatively, users may choose to use previously uploaded data and skip this step. We name this API as \pyth{data_uploading}.
    \item \textbf{Model Listing API\footnote{https://learn.microsoft.com/en-us/rest/api/cognitiveservices/azureopenaistable/models}} The model list API will provide a list of all models that are accessible through the Azure OpenAI resource. Each model in the list includes its name and capabilities, such as whether it supports completion, inference, and fine-tuning. We name this API as \pyth{model_listing}.
    \item \textbf{Fine-tuning API\footnote{https://learn.microsoft.com/en-us/rest/api/cognitiveservices/azureopenaistable/fine-tunes/create}} The fine-tuning API can create a new fine-tuning job given the training dataset and base model. Users can specify the hyper-parameters of training, such as training epochs, batch size, and learning rate. We name this API as \pyth{fine_tune_model}.
    \item \textbf{Job Analyzing API\textsuperscript{\ref{azure_open_ai}}} The job analyzing API can provide the status of the fine-tuning process. The returned values include various information, such as the training step, training loss, and training sequence accuracy. We name this API as \pyth{job_analyzing}.
    \item \textbf{Model Deployment API \textsuperscript{\ref{azure_open_ai}}} The deployment API can create a model deployment for the fine-tuned model, which can then be used by the user like any other models provided by OpenAI. We name this API as \pyth{model_deployment}.
    \item \textbf{Speech-to-text and Text-to-speech API \footnote{https://learn.microsoft.com/en-us/azure/cognitive-services/speech-service/}} Speech-to-text API can convert audio to text from a range of sources, including microphones, audio files, and blob storage. Text-to-speech API can convert input text into humanlike synthesized speech. We name these APIs as \pyth{speech_to_text, text_to_speech}.
\end{itemize}

In this example, the user provides only high-level instruction. However, with detailed documentation for fine-tuning the model step-by-step, TaskMatrix.AI is able to manage the entire conversation and even can answer user questions like "what should I do first" and "what's next". This illustrates the powerful potential of composing instructions, which teaches TaskMatrix.AI to achieve high-level intents by composing multiple APIs.

\subsection{Robotics and IoT Devices Controlling}
\label{sec:iot}
TaskMatrix.AI can help users to interact with the real world by instructing robots and IoT devices. MCFM can understand the environment with camera API, and transform user instructions to action APIs provided by robots and IoT devices. TaskMatrix.AI can facilitate the handling of physical work with the assistance of robots and the construction of smart homes by connecting IoT devices.

\begin{figure*}
    \centering
\includegraphics[width=1.0\textwidth]{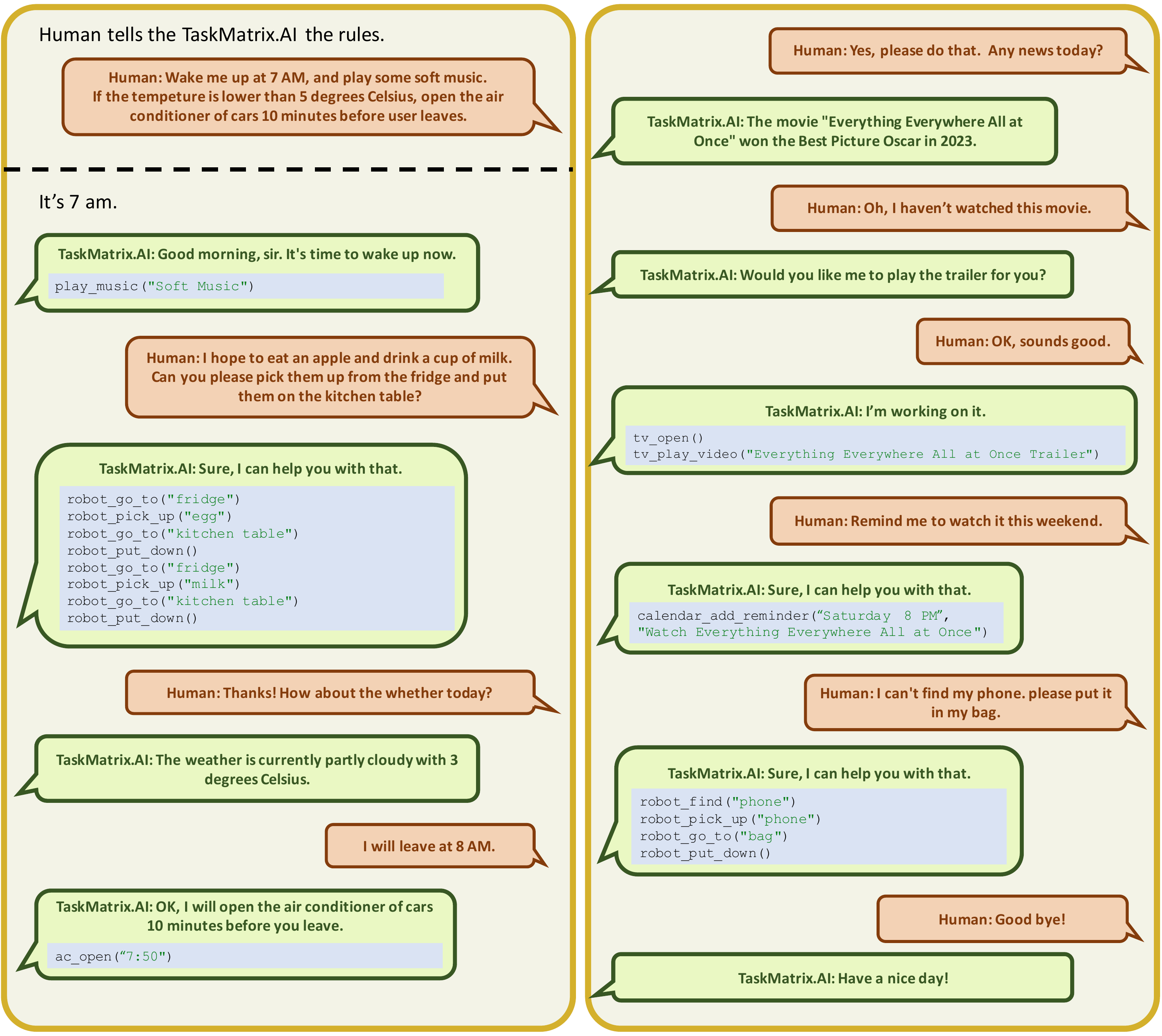} 
\caption{TaskMatrix.AI in smart home scenarios. The user told TaskMatrix.AI several rules before the conversation began, and the conversation is triggered by "it's 7 am". TaskMatrix.AI is capable of integrating with various devices in the user's home, robots, cars, and accessing internet services. By combining these APIs, TaskMatrix.AI can assist users in controlling their devices using voice commands, scheduling items on their calendars, accessing the internet, and taking actions based on specific conditions.} 
\label{fig:smart_home}
\end{figure*}

We show an example in Figure \ref{fig:smart_home}. TaskMatrix.AI can utilize the robotics described in PaLM-E~\citep{driess2023palme} and Microsoft Robotics ~\citep{vemprala2023chatgpt_robotics} to perform tasks such as picking and placing objects, controlling IoT devices in the home. In addition, several popular internet services, such as calendar API, weather API, and news API are included in this scenario. The robotics and IoT devices APIs related to this example include:
\begin{itemize}
    \item \textbf{Robotics Pick, Move and Put APIs} The robots described in PaLM-E and Microsoft Robotics are capable of finding objects, picking up, moving around, and placing objects using their robotic arms, we name them as \pyth{robot_find, robot_pick_up, robot_put_down, robot_go_to}. Although this example utilizes only four APIs, it's worth noting that robots have the potential to support many more APIs in the future.
    \item \textbf{Car Air Conditioner API} The car APIs allow for the control of various car devices, including the air conditioner, location services, and others. The air conditioner API enables users to remotely turn the air conditioner on or off and adjust the temperature settings. We name the APIs as \pyth{ac_open, ac_close, ac_set_temperature}.
    \item \textbf{TV API} The TV API enables users to remotely control their televisions by opening or closing them and playing specific videos. We name the APIs as \pyth{tv_open, tv_close, tv_play_video}. Most smart TVs support this function.
    \item \textbf{Music playing API} The music playing API is capable of playing music on speakers and can receive parameters to specify the name or style of the desired song. We name the API as \pyth{play_music}.
\end{itemize}

In this example, TaskMatrix.AI is able to connect user instructions to APIs of robots and IoT devices. We have pre-defined rules, such as the rule for activating the car's air conditioner which requires two conditions to be met: the temperature must be below 5 degrees Celsius and the user must be leaving. TaskMatrix.AI can accurately execute the API call when both conditions are satisfied. Additionally, the model is capable of correctly filling the parameter with the time "7:50", which is ten minutes before the user's departure.

\subsection{More Scenarios}
\label{sec:more}
TaskMatrix.AI can be applied to much more scenarios by connecting with their APIs. Here, we list three scenarios as examples:

\textbf{Accessing Internet} TaskMatrix.AI can assist users in accessing knowledge and services on the internet. For example, New Bing has successfully leveraged ChatGPT to generate search keywords and summarize search results. TaskMatrix.AI also can interact with internet services such as planning traveling, booking flights, finding products, and replying to emails. This has the potential to facilitate the development of the next-generation web browser and voice assistant.

\textbf{Accessing Metaverse} The Metaverse includes a blend of digital and physical worlds, and TaskMatrix.AI can help users access it in the same way they access operating systems and the internet in digital worlds, as well as robots and IoT devices in physical worlds. Additionally, it can create new simulation experiences by conversing with and instructing AI agents. For instance, Deepmind has developed various AIs for games and virtual environments, such as a team of AIs to play football fully automatically ~\citep{liu2022deepmind_football}. By providing several high-level APIs and integrating with TaskMatrix.AI to build a natural language interface, human players can ask their AI teammates to execute specific tactics and collaborate with them, making the player feel like a team leader. This would enhance the fun factor, as players would no longer be limited to controlling only themselves or letting AI control all units in Deepmind scenarios.

\textbf{Achieving Neuro-Symbolic AI} TaskMatrix.AI can achieve neuro-symbolic integration by formulating user instructions and accessing symbolic modules as API. The symbolic modules can include formal reasoning engines, algorithms implemented by programming languages, and software like Matlab and Mathematica. The expert systems built by humans have been well-verified and exhibit good consistency in solving their targeted problems. By delegating appropriate tasks to expert systems rather than having large models handle everything, TaskMatrix.AI can significantly improve the quality of its output.
\section{Case Study}
\label{sec:case_study}
To demonstrate how TaskMatrix.AI operates, we will take PowerPoint automation as a case study. We aim to use general methods for each module so that the whole pipeline can be easily adapted to other situations. We will explain how we implement MCFM, the API platform, the action executor, and the feedback to API developers in the next subsections. We skip the API selector, because we only need a few PowerPoint APIs for this case. We also postpone RLHF for future work, since we do not have enough user feedback at this point.

\subsection{Multimodal Conversational Foundation Model}
We utilize ChatGPT\footnote{https://learn.microsoft.com/en-us/azure/cognitive-services/openai/how-to/chatgpt} as the MCFM in this scenario. The inputs to the MCFM include the API platform, conversational context, and general prompts. The general prompts aim to align the MCFM with the needs of TaskMatrix.AI, by instructing the model to follow user instructions, generate action codes using defined APIs, and specify the output format.

The PowerPoint content includes the text, position of each text box, images, and other shapes on each slide. With textual content, the MCFM can process textual queries such as "insert the corresponding company logo onto each slide" without requiring specific company names in user instructions. With visual content, the MCFM is capable of resizing and moving the image based on its current size and position.

Since ChatGPT cannot process images, we utilize PowerPoint APIs to parse the visual content into structured textual data. To extend this capability to other software, we can access the operating system APIs or employ tools like ScreenParser ~\citep{wu2021screenparser} to parse elements on the screen. In the future, we can leverage GPT-4 to directly comprehend image inputs. The PowerPoint API we use is the python-pptx package\footnote{https://python-pptx.readthedocs.io/}, which provides the textual content and relative position of each text box and image. We employ a light rectangle detection model to identify the area that displays the current slide. Then, we calculate the position of each shape on the screen and leverage mouse APIs to move and resize objects, similar to how humans interact with the interface. An example of parsed PPT content is shown in Figure \ref{fig:ppt_content_reader}.


\begin{tcolorbox}[enhanced,sharp corners,colback=white,colframe=black,width=\linewidth,boxrule=0.7pt,grow to left by=0.5mm,grow to right by=2.5mm,left=0.5mm,bottom=1mm,attach boxed title to bottom center,coltitle=black,boxed title style={colframe=white,colback=white,width=\linewidth}]
Page: 3\\
Title: Microsoft\\
Contents:\\
Founder: Bill Gates and Paul Allen\\
Location: Redmond, Washington\\
Mission: To empower every person and every organization on the planet to achieve more\\
Products: Windows, Office, Xbox, Bing, Skype, Surface, HoloLens, Dynamics 365\\
Subsidiaries: LinkedIn, GitHub, Mojang, Skype Technologies, Yammer\\

Visual Positions:\\
A text box of title, height=1325, width=10515, it's position: left=838, top=365,right=11353, down=1690\\
A text box of content, height=4351, width=10515, it's position: left=838, top=1825,right=11353, down=6176\\
A picture, height=4572, width=4572, it's position: left=3810, top=1143,right=8382, down=5715\\
\end{tcolorbox}

\begin{center}
\captionof{figure}{The structured content of one PowerPoint page, which contains textual content and the relative position of each text box and image.}
\vspace{0.5cm}
\label{fig:ppt_content_reader}
\end{center}
\definecolor{responsecolor}{gray}{0.85}

\subsection{API Platform}
\label{sec:case_study_api_platform}
This subsection demonstrates the construction of the API platform for PowerPoint and how to teach TaskMatrix.AI to utilize these APIs. Previous research ~\citep{vemprala2023chatgpt_robotics, wu2023visualchatgpt} has demonstrated the importance of API names, descriptions, and parameter lists in enabling correct API usage. In this study, we emphasize the importance of composition instructions for composing multiple APIs to complete complex user instructions. This is demonstrated through subsequent ablation studies.

In Figure \ref{fig:ppt_case1}, we demonstrate how TaskMatrix.AI can be instructed to generate multiple slides, each corresponding to a different company. The API platform for PowerPoint consists of a list of APIs, each accompanied by its name, parameter list, description, and composition instructions, as detailed in Section \ref{sec:overview_api_platform}. These properties are summarized in a single paragraph for ease of understanding. We highlight the composition rules with light green.

\colorlet{mylightgreen}{green!15}
\tcbsetforeverylayer{enhanced,breakable,sharp corners,colback=white,colframe=black,width=\linewidth,boxrule=0.7pt,grow to left by=0.5mm,grow to right by=2.5mm,left=0.5mm,bottom=1mm,attach boxed title to bottom center,coltitle=black,boxed title style={colframe=white,colback=white}}

\newsavebox\lstbox
\begin{lrbox}{\lstbox}\begin{minipage}{0.98\linewidth}
MCFM: Sure, here's the code to generate slides for each company:
\begin{pythongray}
create_slide();
select_title();
insert_text("Microsoft");
select_content();
insert_text("Founder: Bill Gates and Paul Allen\nLocation: Redmond, Washington\nMission: To empower every person and every organization on the planet to achieve more\nProducts: Windows, Office, Xbox, Bing, Skype, Surface, HoloLens, Dynamics 365\nSubsidiaries: LinkedIn, GitHub, Mojang, Skype Technologies, Yammer");
create_slide();
select_title();
insert_text("Apple");
select_content();
insert_text("Founder: Steve Jobs, Steve Wozniak, and Ronald Wayne\nLocation: Cupertino, California\nMission: To design, develop, and sell consumer electronics, computer software, and online services\nProducts: iPhone, iPad, Mac, Apple Watch, Apple TV, HomePod, AirPods, iPod, Apple Music, Apple Pay, iCloud\nSubsidiaries: Beats Electronics, FileMaker Inc., Shazam Entertainment Ltd.");
...
\end{pythongray}
\end{minipage}\end{lrbox} 

\begin{tcolorbox}

\begin{tcolorbox}[enhanced,colframe=black!30,title=API Documentations,attach boxed title to top left={xshift=0.5cm,yshift=-\tcboxedtitleheight/2},grow to right by=1mm,bottom=5mm] 
As an assistant for generating and editing slides, you have access to a list of APIs to control PowerPoint with the following functions:\\
\pyth{create_slide()}: This API is used to create a new slide.\\
\pyth{insert_text(text:str)}: This API is used to insert text into a text box. \colorbox{mylightgreen}{The content of each} \colorbox{mylightgreen}{slide should contain several sentences, you can call \pyth{insert_text} multiple times, or split the} \colorbox{mylightgreen}{multiple sentences by '\textbackslash n'.}\\
\pyth{select_title()}: This API is used to select the text box of the title. \colorbox{mylightgreen}{You should first select} \colorbox{mylightgreen}{the text box of the title and then insert or delete the text in the text box of the title.}\\
\pyth{select_content()}: This API is used to select the text box of the content. \colorbox{mylightgreen}{You should first} \colorbox{mylightgreen}{select the text box of the content and then insert or delete the text in the text box of the content.}\\
\pyth{move_to_slide(slide_id:int)}: This API is used to move to a specific slide in the presentation. It can take one parameter, \pyth{slide_id}: the ID of the slide to move to as an integer. \\
...
\end{tcolorbox}

\begin{tcolorbox}[enhanced,breakable,colframe=black!30,title=Conversational Context,attach boxed title to top left={xshift=0.5cm,yshift=-\tcboxedtitleheight/2},grow to right by=1mm] 
The current version of PPT is:\\
Page: 1\\
Title: Big Technical Companies\\
Visual Positions:\\
...\\

Page: 2\\
Title: Big Technical Companies\\
Contents:\\
Microsoft (1975)\\
Apple (1976)\\
Amazon (1994)\\
Google (1998)\\
Facebook (2004)\\
Visual Positions:\\
...\\

The History of our conversation: \\
Human: I hope to create a PPT about big technical companies. Can you create a slide to list some of them? \\
AI: ... 
\end{tcolorbox}
\begin{tcolorbox}[enhanced,colframe=black!30,title=General Prompts,attach boxed title to top left={xshift=0.5cm,yshift=-\tcboxedtitleheight/2},grow to right by=1mm]   
Don't define new functions.\\
In the output, each line should contain only one function. And each line should end with a ";".\\
Please finish my following instruction with the functions I defined.
\end{tcolorbox}

Human: For each company, let’s create one slide to introduce its founder, location, mission, products, subsidiaries:

\vspace{0.2cm}
\noindent\fcolorbox{white}{gray!15}{\usebox\lstbox}

\end{tcolorbox}

\begin{center}
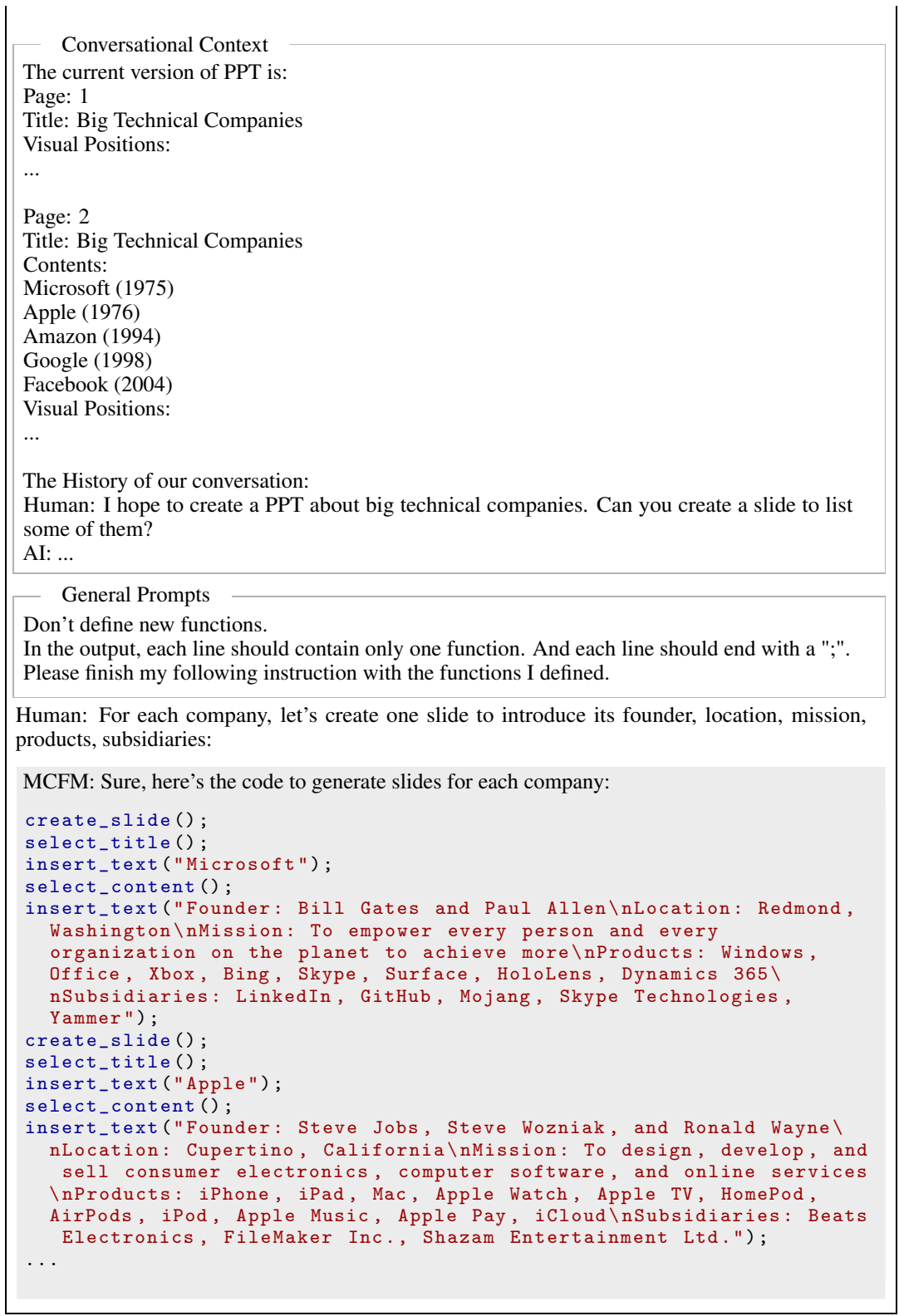

\captionof{figure}{MCFM takes user instruction to generate multiple slides. The inputs to MCFM include API documentation, conversation content, general prompts, and human instructions, all of which are displayed with a white background. Meanwhile, MCFM's outputs are shown with a gray background. }
\label{fig:ppt_case1}
\end{center}

In this case, the API documentation provides the necessary APIs to accomplish this task. The user doesn't provide the company list but it can be obtained from the content of PowerPoint. Since this is a complex instruction, TaskMatrix.AI must break it down into roughly 25 API calls to complete the task. We omit the step of generating a solution outline, as ChatGPT can accurately generate it directly.

The process of decomposing user instructions may vary depending on the API design being used. Thus, it is essential for API developers to provide composition instructions to guide API usage. We have created composition instructions for three APIs: \pyth{insert_text}, \pyth{select_title}, and \pyth{select_content}. The composition instructions for \pyth{insert_text} specify that the content often contains multiple sentences. For \pyth{select_title} and \pyth{select_content}, the composition instructions specify the order in which these APIs should be used with other APIs, such as inserting and deleting text in a text box. These composition instructions can cover multiple combination examples in a single sentence, making them more efficient than in-content learning, which would require multiple examples to cover all these combinations.

\newsavebox\lstboxnoorder
\begin{lrbox}{\lstboxnoorder}\begin{minipage}{0.98\linewidth}
MCFM: Sure, here's the code to generate slides for each company:
\begin{pythongray}
create_slide();
insert_text("Microsoft");
select_title();
insert_text("Founder: Bill Gates, Location: Redmond, Washington, Mission: To empower every person and every organization on the planet to achieve more, Products: Windows, Office, Xbox, Subsidiaries: LinkedIn, Skype");
select_content();
create_slide();
insert_text("Apple");
select_title();
insert_text("Founder: Steve Jobs and Steve Wozniak, Location: Cupertino, California, Mission: To bring the best user experience to its customers through its innovative hardware, software, and services, Products: iPhone, iPad, Mac, Apple Watch, Subsidiaries: Beats Electronics, Shazam");
select_content();
...
\end{pythongray}
\end{minipage}\end{lrbox} 


\begin{tcolorbox}
\begin{tcolorbox}[enhanced,colframe=black!30,title=API Documentations,attach boxed title to top left={xshift=0.5cm,yshift=-\tcboxedtitleheight/2},grow to right by=1mm] 
As an assistant for generating and editing slides, you have access to a list of APIs to control PowerPoint with the following functions:\\
\pyth{create_slide()}: This API is used to create a new slide.\\
\pyth{insert_text(text:str)}: This API is used to insert text into a text box. \colorbox{mylightgreen}{\st{The content of each}} \colorbox{mylightgreen}{\st{slide should contain several sentences, you can call insert\_text multiple times, or split the}} \colorbox{mylightgreen}{\st{multiple sentences by '\textbackslash n'.}}\\
\pyth{select_title()}: This API is used to select the text box of the title. \colorbox{mylightgreen}{\st{You should first select}} \colorbox{mylightgreen}{\st{the text box of the title and then insert or delete the text in the text box of the title.}}\\
\pyth{select_content()}: This API is used to select the text box of the content. \colorbox{mylightgreen}{\st{You should first}} \colorbox{mylightgreen}{\st{select the text box of the content and then insert or delete the text in the text box of the content.}}\\
\pyth{move_to_slide(slide_id:int)}: This API is used to move to a specific slide in the presentation. It can take one parameter, \pyth{slide_id}: the ID of the slide to move to as an integer. \\
...
\end{tcolorbox}
...
\vspace{0.2cm}

\fcolorbox{white}{gray!15}{\usebox\lstboxnoorder}
\end{tcolorbox}

\begin{center}
\captionof{figure}{When composition instructions are removed, MCFM may generate APIs in an incorrect order, such as inserting text before selecting the target text box. Additionally, MCFM may generate a long paragraph for each slide without proper line-breaking.}
\label{fig:ppt_case3}
\end{center}

Figure \ref{fig:ppt_case3} illustrates the results obtained by removing all composition instructions. The model produces a lengthy paragraph without any line breaks. Conversely, the results in Figure \ref{fig:ppt_case1} contain content with multiple lines separated by \pyth{"\n"}, and the model calls \pyth{insert_text} before selecting the target text box. In our experiments, we added irrelevant APIs, changing the prompt to MCFM, and found that there is no consistent order for inserting text and selecting the text box. However, including the composition instructions for \pyth{select_title} and \pyth{select_content} ensures that MCFM always selects the target text box before inserting the text.

Throughout our experiments, we also observed that the model was sometimes able to execute user instructions correctly, but the results were not consistent. This inconsistency may be due to the fact that the model encountered similar knowledge during pre-training, especially for long-standing scenarios that existed prior to the foundation model pre-training. However, we strongly encourage API developers to include composition instructions in order to improve stability, particularly when designing new APIs after foundation model pre-training.

\subsection{Action Executor}

\begin{figure}[h]
    \centering
    \includegraphics[width=1.0\textwidth]{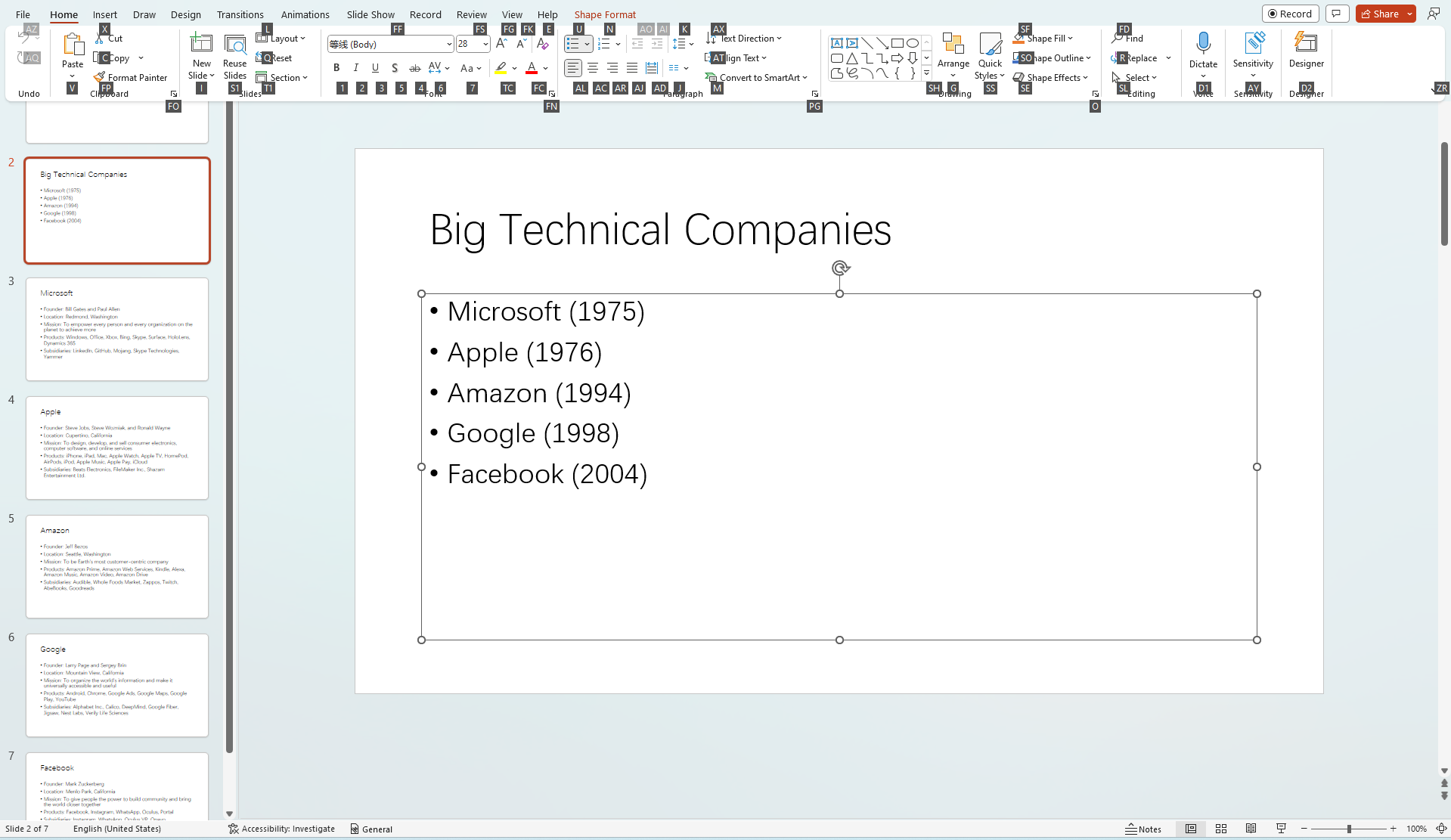}
    \caption{A screenshot of PowerPoint software. When using PowerPoint software, you can press the "alt" key to display one or several characters on each button. This allows you to select the corresponding button using the keyboard.}
    \label{fig:keyboard_for_ppt}
\end{figure}

We utilized the mouse and keyboard API to execute the action codes, as it's a universal method for manipulating the operating system. Specifically, in our experiments, we chose to use the APIs provided in the PyAutoGUI package\footnote{https://pyautogui.readthedocs.io/} for Python. We used the keyboard API to control functions in PowerPoint that have keyboard shortcuts available. When the user clicks the Alt key, PowerPoint displays a shortcut instruction for each button, enabling the user to select the corresponding button using the keyboard. An example is shown in Figure \ref{fig:keyboard_for_ppt}. We also used the mouse API to resize and move images and shapes. The pixel position to operate on was calculated based on the detected position of the area that shows the current slide and the related position of each element provided by the PowerPoint API.

\subsection{Feedback to API Developers}
\label{sec:feedback_to_api_doc}
API developers can modify the API documentation based on user feedback, including cases where the API fails to execute user instructions. Such cases can provide valuable information for improving API documentation. In this work, we demonstrated that a powerful foundation model, such as ChatGPT, can induce general knowledge in natural language based on specific cases and automatically improve API documentation. An example is shown in Figure \ref{fig:ppt_case4}.
\newsavebox\lstboxfeedback
\begin{lrbox}{\lstboxfeedback}\begin{minipage}{0.98\linewidth}
\begin{tcolorbox}[enhanced,colframe=black!30,title=Wrong API calls,attach boxed title to top left={xshift=0.5cm,yshift=-\tcboxedtitleheight/2},grow to right by=1mm] 
Here is the wrong output:
\begin{python}
create_slide();
insert_text("Microsoft");
select_title();
insert_text("Founder: Bill Gates, Location: Redmond, Washington, Mission: To empower every person and every organization on the planet to achieve more, Products: Windows, Office, Xbox, Subsidiaries: LinkedIn, Skype");
select_content();
\end{python}
\end{tcolorbox}
\begin{tcolorbox}[enhanced,colframe=black!30,title=Correct API calls,attach boxed title to top left={xshift=0.5cm,yshift=-\tcboxedtitleheight/2},grow to right by=1mm]
Here is the correct output:
\begin{python}
create_slide();
select_title();
insert_text("Microsoft");
select_content();
insert_text("Founder: Bill Gates and Paul Allen\nLocation: Redmond, Washington\nMission: To empower every person and every organization on the planet to achieve more\nProducts: Windows, Office, Xbox, Bing, Skype, Surface, HoloLens, Dynamics 365\nSubsidiaries: LinkedIn, GitHub, Mojang, Skype Technologies, Yammer");
\end{python}
\end{tcolorbox}
\end{minipage}\end{lrbox} 

\begin{tcolorbox}

\begin{tcolorbox}[enhanced,colframe=black!30,title=API Documentations,attach boxed title to top left={xshift=0.5cm,yshift=-\tcboxedtitleheight/2},grow to right by=1mm] 
As an assistant for generating and editing slides, you have access to a list of APIs to control PowerPoint with the following functions:\\
\pyth{create_slide()}: This API is used to create a new slide.\\
\pyth{insert_text(text:str)}: This API is used to insert text into a text box. \colorbox{mylightgreen}{\st{The content of each}} \colorbox{mylightgreen}{\st{slide should contain several sentences, you can call insert\_text multiple times, or split the}} \colorbox{mylightgreen}{\st{multiple sentences by '\textbackslash n'.}}\\
\pyth{select_title()}: This API is used to select the text box of the title. \colorbox{mylightgreen}{\st{You should first select}} \colorbox{mylightgreen}{\st{the text box of the title and then insert or delete the text in the text box of the title.}}\\
\pyth{select_content()}: This API is used to select the text box of the content. \colorbox{mylightgreen}{\st{You should first}} \colorbox{mylightgreen}{\st{select the text box of the content and then insert or delete the text in the text box of the content.}}\\
...
\end{tcolorbox}

\begin{tcolorbox}[enhanced,colframe=black!30,title=Conversational Context,attach boxed title to top left={xshift=0.5cm,yshift=-\tcboxedtitleheight/2},grow to right by=1mm] 
The current version of PPT are: \\
...\\
The History of our conversation:\\
...\\
\end{tcolorbox}

Human: For each company, let’s create one slide to introduce its founder, location, mission, products, subsidiaries.\\


\begin{tcolorbox}[enhanced,colframe=black!30,title=Wrong API calls,attach boxed title to top left={xshift=0.5cm,yshift=-\tcboxedtitleheight/2},grow to right by=1mm] 
Here is the wrong output:
\begin{python}
create_slide();
insert_text("Microsoft");
select_title();
insert_text("Founder: Bill Gates, Location: Redmond, Washington, Mission: To empower every person and every organization on the planet to achieve more, Products: Windows, Office, Xbox, Subsidiaries: LinkedIn, Skype");
select_content();
\end{python}
\end{tcolorbox}
\begin{tcolorbox}[enhanced,colframe=black!30,title=Correct API calls,attach boxed title to top left={xshift=0.5cm,yshift=-\tcboxedtitleheight/2},grow to right by=1mm]
Here is the correct output:
\begin{python}
create_slide();
select_title();
insert_text("Microsoft");
select_content();
insert_text("Founder: Bill Gates and Paul Allen\nLocation: Redmond, Washington\nMission: To empower every person and every organization on the planet to achieve more\nProducts: Windows, Office, Xbox, Bing, Skype, Surface, HoloLens, Dynamics 365\nSubsidiaries: LinkedIn, GitHub, Mojang, Skype Technologies, Yammer");
\end{python}
\end{tcolorbox}

Please compare the wrong output and correct and summarize your experiences. \\
Your experiences should be instructions that can help to use the APIs you have.\\
List your experiences in bullets.\\

\vspace{0.2cm}

\fcolorbox{white}{gray!15}{
\parbox{0.97\textwidth}{
MCFM: Based on the given instructions and examples, my experiences using the provided APIs are: 
\begin{itemize}
    \item It is important to select the appropriate text box (title or content) before inserting text into it. 
    \item It is necessary to properly format the text (using "\textbackslash n" for line breaks, etc.) before inserting it into the text box.
    \item It is important to provide complete information (founder, location, mission, products, subsidiaries) for each company.
    \item It is helpful to use a consistent format for presenting information about each company.
    \item It is important to create a new slide for each company.
\end{itemize}
}
}
\end{tcolorbox}

\begin{center}
\captionof{figure}{ChatGPT generates feedback to API developers given the <user instruction, wrong API calls, correct API calls>. We can see that the first two items of feedback can be helpful in improving API documentation.}
\label{fig:ppt_case4}
\end{center}

Specifically, we utilized ChatGPT with inputs consisting of <user instruction, wrong API calls, correct API calls>, where the correct API calls were obtained from user demonstrations. We provided the API documentation without any composition instructions and instructed the foundation models to summarize the feedback on using APIs. Based on the results, we found that the first two items of feedback were highly related to our composition instructions, while the other feedback items were specific to this particular case. API developers can select the first two feedback items and add them to the API documentation. This process can also be automated by incorporating a separate module to test each feedback item individually and adding them when the test results demonstrate improvement.
\section{Challenges} \label{sec:exp}
Although we have conducted preliminary validation of tasks in several scenarios, there are still many challenges to fully realizing TaskMatrix.AI's vision.

\textbf{Multimodal Conversational Foundation Model} To handle various tasks, TaskMatrix.AI needs a powerful foundation model that can work with different kinds of inputs (such as text, image, video, audio, code, etc.), learn from context, use common sense to reason and plan, and generate high-quality codes based on APIs to complete tasks. ChatGPT can only deal with text and code and is not able to handle tasks that involve physical and multimodal inputs. GPT-4~\citep{openai2023gpt4} is a better choice as it can process multimodal inputs (i.e., text and image). However, TaskMatrix.AI also needs to handle other modalities that may be returned by different APIs besides text, code and image. It is still challenging to figure out the minimum set of modalities that TaskMatrix.AI requires and train an MCFM for it.

\textbf{API Platform} Creating and maintaining a platform that hosts millions of APIs requires solving several challenges, such as: 1) API documentation generation: In the early stage, most APIs lack proper documentations that are friendly for MCFM to understand and invoke. 2) API quality assurance: The quality and dependability of APIs can differ greatly. It is crucial to ensure that the APIs in the platform meet the necessary quality criteria and are trustworthy for the success of TaskMatrix.AI. 3) API creation suggestion: Based on the user feedback, TaskMatrix.AI can recognize the shortcomings of existing APIs and know which kinds of tasks they cannot handle. Based on these, the API platform should provide further guidance for API developers to create new APIs to address such tasks.

\textbf{API Calling} Leveraging millions of APIs to accomplish user instructions raise new challenges that go beyond free text generation:  1) API selection: When there are many APIs available, recommending related APIs to MCFM for solving a specific task is vital. It requires TaskMatrix.AI to have a strong ability to plan reasonable solutions that can link user intentions with suitable APIs based on their documentation and previous usage history. 2) Online planning: For complex tasks, TaskMatrix.AI may not be able to come up with a solution right away. Instead, MCFM should interact with users and try different possible solutions to figure out the most suitable one. 

\textbf{Security and Privacy} When the API can access the physical world, and digital world to make real changes, we need to make sure the model: 1) Faithful to user instructions. We need to verify that the model accomplishes user instructions and doesn't do anything more than the user's intent. 2) Keep data private. The data transmission should be secure and data access should be authorized when integrating with various APIs from different domains that require access to sensitive data.

\textbf{Personalization} TaskMatrix.AI requires a personalization strategy to assist individual developers in building their own personalized AI interfaces for their products, as well as to help users have their own personal assistants. Personalization faces two challenges: 1) Reducing scaling cost. Since the learning needs to apply to numerous developers and users, it is not feasible to fine-tune one model for each scenario. 2) alignment with the user with few-shot examples. As users may only provide a few demonstrations or feedback and the model needs to efficiently learn their preferences. One promising direction is to generate a preference-aware solution outline.

\section{Related Work}

Improving the performance of specific tasks with existing APIs has been studied in various scenarios. For example, WebGPT~\citep{nakano2021webgpt}, ReAct~\citep{yao2022react}, and ~\citet{ lazaridou2022internet} leveraged search APIs to provide comprehensive information for more trustworthy text generation. ChatGPT Robotics~\citep{vemprala2023chatgpt_robotics}, PaLM-SAYCAN~\citep{ahn2022saycan}, PaLM-E~\citep{driess2023palme} and \citet{liang2022code_as_policy} instructed robotics to finish physical world tasks by leveraging high-level robotics APIs. To solve mathematical problems, ~\citet{cobbe2021gsm8k} used a calculator to fix the calculation error, ~\citet{gao2022pal} used the code interpreter to execute the code generated from the input text, ~\citet{jiang2022draft_sketch_prove} leverage mathematical prover to prove the complex mathematical theory. ToolFormer~\citep{schick2023toolformer} leveraged search API, question answering API, machine translation API and calculator to solve various NLP tasks. ART~\citep{paranjape2023art} leveraged five kinds of tools (arithmetic, code, search, free-form reasoning and string operations) to improve the performance on BigBench \citep{ghazal2013bigbench}. ~\citet{mialon2023augmentedlm} provided a detailed survey for these works. Visual ChatGPT ~\citep{wu2023visualchatgpt} and MM-REACT~\citep{yang2023mmreact} incorporated multiple visual models for better image generation and understanding. However, all these works only focus on a few fixed APIs in a specific domain. We propose to build an API platform that may contain millions of APIs to help solve problems in various domains.

There are four different methods for teaching models using APIs. First, Galactica~\citep{taylor2022galactica} pre-trained the model and ToolFormer~\citep{schick2023toolformer} fine-tuned the model with a corpus of examples that utilize APIs. Second, create a few examples that use APIs and leverage in-context learning to teach the model ~\citep{saycan2022arxiv, gao2022pal, lazaridou2022internet}. Third, leverage reinforcement learning with human feedback to improve the performance of use APIs ~\citep{nakano2021webgpt}. Fourth,  create natural language documents ~\citep{vemprala2023chatgpt_robotics} or structure programs ~\citep{paranjape2023art} to instruct the model on how to use APIs. The pre-training and fine-tuning approach requires a fixed API space and is hard to adapt to API updates. And in-context learning can't handle a large number of APIs. We leverage API documents to connect user instructions to APIs, and we leverage RLHF to improve the connection ability from user feedback. We also implement a feedback loop to assist API developers in improving their documentation, which can achieve lifelong learning with natural language.

Several preliminary products share a similar idea of TaskMatrix.AI, such as the ACT-1 of ADEPT\footnote{https://www.adept.ai/blog/act-1}, which target building models that can take actions in the digital world. LangChain\footnote{https://python.langchain.com/en/latest/index.html} targets to combine multiple language models and other sources of computation or knowledge that take in a string and return a string. We also proposed and open-sourced Visual ChatGPT\footnote{https://github.com/microsoft/visual-chatgpt} as the first scenarios of TaskMatrix.AI, which can handle complex visual tasks, such as image-based question answering, generation, and editing. The ChatGPT Plugins \footnote{https://openai.com/blog/chatgpt-plugins} of OpenAI can help ChatGPT access up-to-date information, run computations, or use third-party services. Together with these works, we aim to facilitate research in this area and share ideas about how to implement similar products.
\section{Looking Forward}

TaskMatrix.AI is a platform that allows people to perform diversified tasks by connecting foundation models with various existing systems and models via their APIs. With the fast development of the foundation model, cloud service, robotics, and Internet of Things technologies and infrastructures, we can imagine an amazing future world, where productivity and creativity can reach new levels.

\bibliography{custom}
\bibliographystyle{nips}

\end{document}